\documentclass[runningheads]{llncs}
\usepackage[T1]{fontenc}
\usepackage{graphicx}
\usepackage{booktabs}

\usepackage{hyperref}
\usepackage{bookmark}
\hypersetup{hidelinks}
\bookmarksetup{open,numbered,depth=2}

\usepackage[misc]{ifsym}
\newcommand{\corr}{(\Letter)}
\usepackage{amsmath,amssymb}
\usepackage{microtype}
\usepackage{multirow}
\usepackage{mwe}
\usepackage{tabularx}
\usepackage{placeins} 
\newcolumntype{Y}{>{\centering\arraybackslash}X}
\newcolumntype{R}{>{\raggedleft\arraybackslash}X}
\newcolumntype{Z}{>{\raggedleft\arraybackslash}X}

\begin{document}

\title{ALINC: Active Learning for Inductive Node Classification via Graph Sampling}

\titlerunning{Active Learning for Inductive Node Classification via Graph Sampling}

\author{Pascal Plettenberg\inst{1} \corr \and
Denis Huseljic\inst{1} \and
André Alcalde\inst{2} \and
Bernhard Sick\inst{1} \and
Josephine M. Thomas\inst{3}}

\authorrunning{P. Plettenberg et al.}

\institute{Intelligent Embedded Systems, University of Kassel, 34121 Kassel, Germany \email{\{plettenberg, dhuseljic, bsick\}@uni-kassel.de}
\and
CELUS GmbH, 80339 Munich, Germany 
\email{andre.alcalde@celus.io}
\and
GAIN Group, Institute of Data Science, University of Greifswald, 17489 Greifswald, Germany \email{thomasj@uni-greifswald.de}}

\maketitle              

\begin{abstract}
Active learning (AL) for node classification typically focuses on selecting the most informative nodes for annotation within one or a few large graphs (e.g., in social network analysis). However, in other domains, such as molecular chemistry or electronic design automation, datasets consist of thousands of independent graphs. In many of these inductive settings, annotating an individual node requires a full-graph analysis, which effectively yields the remaining node labels on-the-fly. Therefore, these scenarios require AL strategies that select entire graphs instead of single nodes, a problem which has not been tackled in the literature so far. Thus, we introduce ALINC, an AL framework for inductive node classification via graph sampling. It bridges the existing methodological gap by elevating node-level utility measures to graph-level selection criteria through various aggregation mechanisms. In an extensive benchmark including ten strategies, three aggregation methods, and four datasets, we identify CoreSet, TypiClust, and BADGE as the top-performing graph sampling strategies. Our detailed analysis further reveals that the choice of the aggregation method is pivotal, as it substantially affects model performance and annotation costs. Finally, we demonstrate the effectiveness of ALINC in two use case studies: site-of-metabolism prediction in molecules and design automation of printed circuit board schematics.

\end{abstract}

\section{Introduction}

The remarkable success of graph neural networks (GNNs) across diverse domains is fundamentally tied to the availability of high-quality, labeled datasets \cite{dwivedi2023benchmarking,hu2020open}. However, acquiring these labels is often a time-consuming and costly process that requires domain expertise. To mitigate this, active learning (AL) has emerged as a powerful paradigm~\cite{settles2009active}. By iteratively identifying and selecting the most informative instances for annotation, AL aims to maximize model performance while minimizing the total labeling cost. AL has been extensively studied in traditional data domains such as text~\cite{rauch2023activeglae} and images~\cite{huseljic2024fast,huseljic2025cleaning}, but while established AL methods can be directly transferred to graph-structured data for graph-level prediction tasks~\cite{xie2022active}, their application to node-level tasks is more challenging because inter-node dependencies violate the standard i.i.d.~assumption.

So far, AL for node classification has predominantly focused on the \emph{transductive setting}~\cite{cai2017active,gao2018active,hu2020graph,ma2023partition}, where the dataset usually consists of a single large graph and the entire graph structure is known during training. In this setting, the goal is to select individual nodes for annotation based on different measures of node utility, e.g., uncertainty or representativeness. Only a few works also apply AL to node classification in the \emph{inductive setting}, where the model needs to generalize to unseen graph structures~\cite{song2023no,zhang2022information,zhang2021grain}. However, these studies focus on selecting individual nodes from datasets consisting of a few large-scale graphs, e.g., different snapshots of a growing social network such as in the Reddit dataset~\cite{hamilton2017inductive}.

A critical gap remains in settings where inductive node classification is performed on large datasets composed of thousands of smaller, independent graphs, e.g.,~in molecular chemistry~\cite{reiser2022graph} or electronic design automation~\cite{plettenberg2025graph}. In these contexts, labeling a single node typically requires a holistic assessment of the whole graph (or large parts of it), possibly via expensive experiments or simulations, which often yields many or all of the remaining node labels on-the-fly. In some cases, such as site-of-metabolism (SoM) identification in molecules~\cite{sicho2019fame}, isolated node labeling is even technically infeasible because experimental protocols inherently characterize the entire molecule. Consequently, all node (or atom) labels are generated simultaneously, making the cost of annotating a single node effectively identical to that of the entire graph. In these situations, AL strategies therefore need to select entire graphs for annotation rather than single nodes, which is more challenging because it involves the aggregation of node-level utility scores. To the best of our knowledge, no existing study has so far investigated this integration of aggregation mechanisms into AL strategies to perform graph sampling for inductive node classification.

In this paper, we bridge this gap by introducing an AL framework that enables graph sampling for inductive node classification. Our contributions are the following:
\begin{enumerate}
    \item \textbf{Framework:} We propose \textbf{ALINC}\footnote{The code is available at \href{https://github.com/pasplett/alinc}{https://github.com/pasplett/alinc}.} -- an \textbf{A}ctive \textbf{L}earning framework for \textbf{I}nductive \textbf{N}ode \textbf{C}lassification via graph sampling. ALINC integrates various mechanisms for aggregating node-level utility measures into established AL strategies, including uncertainty-, representativeness-, diversity-based, and hybrid strategies.
    \item \textbf{Benchmarking:} We conduct an extensive benchmark study, comparing ten different strategies and three aggregation methods across four different datasets, using both message-passing GNNs and graph transformers.
    \item \textbf{Analysis:} We provide a detailed analysis of how different combinations of strategies and aggregation methods affect model performance and annotation costs, providing useful insights and recommendations for practitioners.
    \item \textbf{Applicability:} We present two use case studies on real-world graph datasets in the domains of SoM prediction and design automation of printed circuit board (PCB) schematics to highlight the effectiveness of ALINC.
\end{enumerate}

\section{Related Work}

\textbf{GNNs} such as GCN~\cite{kipf2017semi}, GatedGCN~\cite{bresson2017residual,dwivedi2023benchmarking}, GAT \cite{velivckovic2018graph}, GIN~\cite{xu2019powerful}, GINE~\cite{hu2020strategies}, and GATv2 \cite{brody2022attentive} have achieved outstanding results in learning local relationships from graph-structured data via local message-passing. However, their ability to capture long-range dependencies is limited by issues like over-smoothing~\cite{oono2020graph} and over-squashing~\cite{alon2021bottleneck}. To mitigate this, graph transformers (GT)~\cite{chen2022structure,shirzad2023exphormer,ying2021transformers} leverage global attention for long-distance modeling. Recent hybrid architectures, notably GPS~\cite{rampavsek2022recipe}, combine local message-passing with positional encodings and global attention to improve performance. 

\textbf{AL strategies} generally belong to at least one of the following categories: uncertainty, representativeness, and diversity. Traditional uncertainty heuristics, such as entropy and margin~\cite{settles2009active}, are used to prioritize instances near the decision boundary where the model lacks confidence. To ensure the selections are representative of the underlying data distribution, instances can be selected based on information density~\cite{settles2009active}. In batch AL, where multiple instances are sampled at once, diversity-based methods like CoreSet~\cite{sener2018active} treat the data selection as a geometric coverage problem to minimize redundancy across the feature space. Modern hybrid frameworks aim to balance these competing objectives. For instance, BADGE~\cite{ash2020deep} combines gradient embeddings with $k$-\textsc{means++} to identify uncertain yet diverse batches, while TypiClust~\cite{hacohen2022active} integrates clustering with density to select typical instances from across the entire manifold.

\textbf{AL on graphs} has so far focused on strategically selecting the most informative nodes for annotation, and has been mainly applied to transductive node classification~\cite{cai2017active,gao2018active,hu2020graph,ma2023partition} as well as inductive node classification on a single or a few large graphs~\cite{song2023no,zhang2022information,zhang2021grain}. The most fundamental approaches include uncertainty-based strategies \cite{fuchsgruber2024uncertainty}, density-based sampling~\cite{cai2017active}, as well as strategies based on node centrality, e.g., node degree~\cite{cai2017active} or PageRank centrality~\cite{rodriguez2008grammar}, which can be seen as a measure of representativeness. AGE \cite{cai2017active} combines entropy, information density, and node centrality into a single heuristic score. ANRMAB \cite{gao2018active} introduces a dynamic weighting via a multi-armed bandit framework. Other notable strategies for AL on graphs include FeatProp \cite{wu2019active}, which combines node feature propagation with node clustering, GPA \cite{hu2020graph}, which uses reinforcement learning to learn an optimal query strategy, and GraphPart \cite{ma2023partition}, which selects representative nodes from disjoint partitions of the graph. 

However, these strategies are specifically designed for node sampling and are not directly transferable to inductive node classification via graph sampling, which requires the aggregation of node-level measures. We note that this approach is related to \textbf{AL for object detection}, where scores for individual detections are aggregated for image selection \cite{brust2019active,roy2018deep}. To the best of our knowledge, this is the first work to transfer this concept to the graph sampling problem for inductive node classification.

\section{Problem Formulation}\label{sec:problem}

We consider a batch AL setting with a total labeling budget $B$ distributed over $T$ consecutive cycles. Let $\mathcal{U}_t =\{G_1, G_2, \dots, G_N\}$ be a collection of unlabeled graphs at cycle $t$, where each graph $G_i = (\mathcal{V}_i,~\mathbf{A}^{(i)})$ is defined by a set of nodes $\mathcal{V}_i$ and an adjacency matrix $\mathbf{A}^{(i)} \in \{0, 1\}^{|\mathcal{V}_i| \times |\mathcal{V}_i|}$. We assume the existence of an oracle that, when queried with a graph $G_i$, provides a label matrix $\mathbf{Y}^{(i)} \in \mathbb{N}^{|\mathcal{V}_i| \times C}$, i.e., the matrix entry $y^{(i)}_{jk} = 1$ indicates node $j$ has label $k$. The set of labeled graphs $\mathcal{L}_t = \{(G_i, \mathbf{Y}^{(i)})\}$ is initialized at $t=0$ by randomly sampling $b$ graph instances. Afterwards, we select $b$ instances to label in each cycle, such that $B = b + T \cdot b$.

Given a GNN or GT model $f_{\theta} = h_{\omega} \circ g_{\varphi}$ consisting of an encoder $g$ parameterized by $\varphi$ and a classification head $h$ parameterized by $\omega$, the goal of \textbf{active inductive node classification via graph sampling} is to design a strategy that selects in each cycle $t > 0$ a batch of graphs $\mathcal{S}_t \subset \mathcal{U}_t$  with $|\mathcal{S}_t| = b$ that minimizes the expected loss on an unseen test distribution $\mathcal{D}_{test}$:
\begin{equation}
\min_{\mathcal{S}_t \subset \mathcal{U}_t, |\mathcal{S}_t|=b} \mathbb{E}_{G_i \sim \mathcal{D}_{test}} \left[ L(f_{\theta^*}(\{G_i\}), \{\mathbf{Y}^{(i)}\}) \right],
\end{equation}
where $L$ is the cross-entropy loss and the optimal parameters $\theta^* = \{\varphi^*, \omega^*\}$ are obtained by training the model $f_{\theta}$ on the labeled graph set $\mathcal{L}_t = \mathcal{L}_{t-1} \cup \mathcal{S}_t$.

\section{The ALINC Framework}\label{sec:framework}

In the ALINC framework, a query strategy selects in cycle $t$ a subset of graphs $\mathcal{S}_t \subset \mathcal{U}_t$ with $|\mathcal{S}_t| = b$ by ranking the unlabeled graphs in $\mathcal{U}_t$ based on an acquisition function $\Phi: \mathcal{U}_t \rightarrow \mathbb{R}$. Formally, the selection is defined as:
\begin{equation}
    \mathcal{S} = \underset{G_i \in \mathcal{U}_t}{\operatorname{arg\,top-b}} ~~ \Phi(G_i \mid \varphi,\omega,\mathcal{U}_t,\mathcal{L}_{t-1}).
\end{equation}
The acquisition function $\Phi$ evaluates the utility of a graph $G_i$, based on a permutation-invariant aggregation $\text{agg}(\cdot)$ (such as \texttt{max}, \texttt{mean}, or \texttt{sum}) of node-level properties, including latent node embeddings $\mathbf{H}^{(i)} \in \mathbb{R}^{|\mathcal{V}_i| \times M}$ computed by $g_{\varphi}$, predictive logits $\mathbf{Z}^{(i)} \in \mathbb{R}^{|\mathcal{V}_i| \times C}$ computed by $f_{\theta}$, or structural measures derived from the adjacency matrix $\mathbf{A}^{(i)} \in \{0, 1\}^{|\mathcal{V}_i| \times |\mathcal{V}_i|}$. Furthermore, representativeness- and diversity-based strategies are context-aware, as they are also conditioned on all other graphs in $\mathcal{U}_t$ and possibly $\mathcal{L}_{t-1}$.

In the following, we describe how different types of query strategies can be transferred into the ALINC framework. We include both general AL strategies as well as special strategies for AL on graphs, which were originally designed for node sampling.

\subsection{Uncertainty-based Strategies}

Uncertainty-based sampling is a widely used paradigm in AL that identifies instances for which the current model $f_\theta$ lacks confidence. In the ALINC framework, we adapt these strategies by aggregating node-level uncertainty measures to the graph level. For each graph $G_i \in \mathcal{U}_t$, we first compute the predictive probabilities $\mathbf{P}^{(i)} = \sigma(\mathbf{Z}^{(i)})$, where $\sigma(\cdot)$ denotes the softmax function and the matrix entry $p^{(i)}_{jk}$ represents the probability of node $j$ belonging to class $k$. We consider two popular uncertainty metrics~\cite{settles2009active} for the acquisition function $\Phi$:

\textbf{Entropy} sampling selects graphs with the highest aggregated information content. The node-level entropy for node $j$ is defined as $E_j = -\sum_{k=1}^C p^{(i)}_{jk} \log p^{(i)}_{jk} $. The acquisition function is then:
\begin{equation}
    \Phi_{\text{Ent}}(G_i \mid \varphi, \omega) = \text{agg}_{j \in \mathcal{V}_i} (E_j).
\end{equation}
    
\textbf{Margin} sampling targets nodes where the model cannot easily distinguish between the two most likely classes. Let $\hat{k}_1$ and $\hat{k}_2$ be the indices of the largest and second-largest probabilities for node $j$. The margin is $M_j = 
p^{(i)}_{j\hat{k}_1} - p^{(i)}_{j\hat{k}_2}$. Since a smaller margin implies higher uncertainty, we use:
\begin{equation}
    \Phi_{\text{Marg}}(G_i \mid \varphi, \omega) = \text{agg}_{j \in \mathcal{V}_i} (1 - M_j).
\end{equation}

As defined in Section~\ref{sec:framework}, the aggregation operator $\text{agg}(\cdot)$ can be any permutation-invariant function. For example, using \texttt{mean} allows the framework to prioritize graphs with a high overall density of uncertain nodes, while the \texttt{max} operator would prioritize graphs with the highest individual node uncertainty.

\subsection{Representativeness-based Strategies}

Representativeness-based strategies aim to select representative instances from high-density regions of the feature space, giving lower priority to outliers. In graph AL, the most common measures for representativeness are information density \cite{settles2009active} and node centrality \cite{cai2017active}.

\subsubsection{Density-based Strategies.} In conventional AL settings, the density score of an instance $\mathbf{x}_j$ from an arbitrary unlabeled pool $\{\mathbf{x}_j\}$ is computed from the Euclidean distance to its nearest cluster center $\mathbf{c}^*_j$, derived from applying $k$-\textsc{means} to the embedding set:
\begin{equation}
    \text{density}(\mathbf{x}_j \mid \{ \mathbf{x}_j \}) = \frac{1}{1 + \|\mathbf{x}_j - \mathbf{c}^*_j\|_2}.
\end{equation}

Increasing the number of clusters $k$ reduces redundancy in the acquired batch but makes it less representative. We choose $k = b$ in our experiments, following usual practice~\cite{cai2017active,ma2023partition}. However, we provide additional experiments in Appendix~\ref{app:clusters}, showing that $k = b$ is not necessarily the best choice.

In ALINC, we distinguish between two different density-based acquisition strategies. In \textbf{Node-Density} sampling, we first perform the clustering and density calculation on the node embedding $\mathbf{h}^{(i)}_{j}$ from all graphs in $\mathcal{U}_t$. The aggregation is then performed on the node-level density scores:

\begin{equation}
    \Phi_{\text{ND}}(G_i \mid \varphi,\mathcal{U}_t) = \text{agg}_{j \in \mathcal{V}_i}(\text{density}(\mathbf{h}^{(i)}_j \mid \{\mathbf{h}_j^{(i)}\}_{1 \leq i \leq N,~j \in \mathcal{V}_i})).
\end{equation}

In \textbf{Graph-Density} sampling, node embeddings are first aggregated into a single graph-level representation $\mathbf{g}^{(i)} = \text{agg}_{j \in \mathcal{V}_i}(\mathbf{h}^{(i)}_j)$. Clustering and density calculation are then performed directly on the set of graph-level embeddings:

\begin{equation}
    \Phi_{\text{GD}}(G_i \mid \varphi, \mathcal{U}_t) = \text{density}(\mathbf{g}_{i} \mid \{\mathbf{g}^{(i)}\}_{1 \leq i \leq N}).
\end{equation}

While the node-density strategy prioritizes graphs that contain representative nodes with respect to the learned node embedding space, the graph-density strategy prioritizes representative instances on the graph level. 

\subsubsection{Centrality-based Strategies.} The representativeness or importance of a node in a graph can be measured by node centrality scores, such as node degree and PageRank centrality \cite{rodriguez2008grammar}. Centrality-based AL strategies select the nodes with the highest centrality scores. We transfer this approach to the ALINC framework by aggregating node-centrality scores to graph-level utility scores.

Since the PageRank score is a measure of the relative importance of a node in a graph, it is not comparable across different graphs. Therefore, we only consider \textbf{Degree} sampling in the ALINC framework. In \textbf{Degree} sampling, the acquisition function can be formulated as
\begin{equation}
    \Phi_{\text{Deg}}(G_i) = \text{agg}_{j \in \mathcal{V}_i} (d^{(i)}_j)
\end{equation}
with the node degree $d^{(i)}_j = \sum_{k} \mathbf{A}^{(i)}_{jk}$.

\subsection{Diversity-based Strategies}

In batch AL, strictly selecting the top-$b$ instances according to uncertainty or representativeness may introduce redundancy. Therefore, we consider diversity-based strategies to ensure that the selected graphs cover the complete feature space. 

We include \textbf{CoreSet} \cite{sener2018active} as the only purely diversity-based strategy. This strategy aims to select a subset that covers the entire data distribution by minimizing the maximum distance between any unlabeled instance and its nearest labeled neighbor using a k-Center-greedy algorithm. In ALINC, we perform this selection based on the aggregated graph-level representations $\mathbf{g}^{(i)}$.

\subsection{Hybrid Strategies}

We further consider strategies that combine at least two of the three categories: uncertainty, representativeness, and diversity.

\textbf{AGE} \cite{cai2017active} (uncertainty, representativeness) selects nodes for annotation by evaluating the utility of a node via a linear combination of entropy, density, and node centrality converted into percentiles. We transfer AGE to the ALINC framework by utilizing the strategies Entropy, Node-Density, and Degree to compute node-level AGE scores before aggregating them to the graph level afterwards.

\textbf{ANRMAB} \cite{gao2018active} (uncertainty, representativeness) considers the same utility scores as AGE but combines them using a multi-armed bandit (MAB) approach. To perform graph sampling with this strategy, we aggregate the individual scores before applying MAB on the graph level.

\textbf{TypiClust}~\cite{hacohen2022active} (representativeness, diversity) selects typical instances from clusters that are under-represented in the labeled set. Similar to CoreSet, we perform the clustering and density-based selection on the aggregated graph-level embeddings $\mathbf{g}_i$.

Finally, \textbf{BADGE}~\cite{ash2020deep} (uncertainty, diversity) consists of two main computation steps, a gradient embedding computation and a sampling computation. In the first step, the model's current predictions are treated as pseudo-labels for estimating the induced gradient of the loss with respect to the model parameters. In our framework, we aggregate the resulting node-level gradient embeddings to obtain graph-level embeddings. We then proceed with the second step by sampling from the aggregated gradient embeddings via $k$-\textsc{means++}.

\section{Experiments}

In this section, we evaluate the ALINC strategies introduced in Sec.~\ref{sec:framework} on four benchmark graph datasets for inductive node classification. We first describe our experimental setup and then proceed with presenting a comprehensive performance comparison between strategies and aggregation methods. Next, we analyze the annotation costs of the different strategies in terms of the number of nodes in the selected graphs. Finally, we present additional results on two datasets from highly relevant application domains: (i)~Site-of-metabolism (SoM) prediction in molecular graphs and (ii)~printed circuit board (PCB) schematic design automation.

\subsection{Experimental Setup}\label{sec:setup}

\subsubsection{Datasets.} For our benchmark experiments, we use four popular graph datasets for inductive node classification and adopt them in our AL setting: PATTERN and CLUSTER \cite{dwivedi2023benchmarking} contain synthetic graphs with node labels for pattern identification and node clustering, respectively. PascalVOC-SP and COCO-SP are part of the long-range graph benchmark \cite{dwivedi2022long} and contain image-based graphs, where each node corresponds to a superpixel with a particular semantic segmentation label.

\subsubsection{AL Protocol.} We adopt a common batch AL protocol from the image domain~\cite{huseljic2024fast,huseljic2025cleaning} and start from a random initial labeled pool of size $b$ followed by 20 AL cycles with an acquisition size of $b$. Here, $b$ is determined by ensuring that the respective evaluation metric converges within 20 cycles of random sampling. However, we restrict $b$ to a maximum of 100 in order to focus on the typical AL scenario of labeled data scarcity. An overview of all datasets with corresponding batch sizes is given in Tab.~\ref{tab:datasets}.

\begin{table}[!h]
\caption{\label{tab:datasets} Overview of the graph datasets used for benchmarking AL strategies in the ALINC framework.}
\scriptsize
\renewcommand{\arraystretch}{1.5}
\begin{tabularx}{\textwidth}{
    >{\hsize=1.1\hsize}X 
    >{\hsize=0.9\hsize}Z 
    >{\hsize=0.9\hsize}Z 
    >{\hsize=0.9\hsize}Z 
    >{\hsize=1.1\hsize}Z 
    >{\hsize=1.1\hsize}Z 
}
\hline
\textbf{Dataset} & \textbf{\#Graphs} & \textbf{Avg. Nodes} & \textbf{Avg. Edges} & \textbf{\#Classes (C)} & \textbf{Acq. Size (b)} \\
\hline
PATTERN          & 14,000             & 117.47                 & 4,749.15               & 2                   & 5                   \\
CLUSTER          & 12,000             & 117.20                 & 4,301.72               & 6                   & 50                  \\
PascalVOC-SP     & 11,355             & 479.40                 & 2,710.48               & 21                  & 100                 \\
COCO-SP          & 123,286            & 476.88                 & 2,693.67               & 81                  & 100 \\
\hline
\end{tabularx}
\end{table}

\subsubsection{Models.} We use two different models to cover both message-passing GNNs and GTs. For PATTERN and CLUSTER, we use GatedGCN, and for PascalVOC-SP and COCO-SP, we use GPS, since standard message-passing GNNs are significantly outperformed by GTs on the long-range graph benchmark \cite{dwivedi2022long}. 

\subsubsection{Hyperparameters.} The role of hyperparameters in AL is an open problem \cite{huseljic2023role}. In this study, we adopt reported model hyperparameters from the literature, wherever possible. For GatedGCN on PATTERN and CLUSTER, we use the hyperparameters from~\cite{dwivedi2023benchmarking}, and only tune the learning rate by maximizing validation performance on random sampling. For GPS on PascalVOC-SP and COCO-SP, we adopt the hyperparameters from~\cite{tonshoff2023did}. 

\subsubsection{Baseline Strategies} We use random sampling as our baseline and compare it to all strategies introduced in Sec.~\ref{sec:framework}: Entropy \cite{settles2009active}, Margin \cite{settles2009active}, Node-Density, Graph-Density, Degree \cite{cai2017active}, CoreSet \cite{sener2018active}, AGE \cite{cai2017active}, ANRMAB \cite{gao2018active}, TypiClust \cite{hacohen2022active}, and BADGE \cite{ash2020deep}. We further evaluate each strategy with different aggregation mechanisms: \texttt{mean}, \texttt{max}, and \texttt{sum}.

\subsubsection{Evaluation Metrics.} We utilize relative learning curves, which track performance gains in absolute percentage points over random sampling at each cycle (positive values demonstrate AL effectiveness), and the area under the learning curve (AULC) to quantify cumulative performance across all cycles. We further supplement our evaluation with pairwise win rate comparisons, i.e., the percentage of trials where strategy $i$ yields a higher AULC than strategy $j$.

For PATTERN and CLUSTER, performance is measured via macro-averaged accuracy, and for PascalVOC-SP and COCO-SP, the performance corresponds to the macro-averaged F1 score. In all experiments, the performance is measured on a separate test dataset (excluded from the unlabeled pool), averaged across ten independent trials.

\subsection{Benchmarking ALINC Strategies}\label{sec:perf1}

\begin{figure}[t]
\centering
\includegraphics[width=0.8\textwidth]{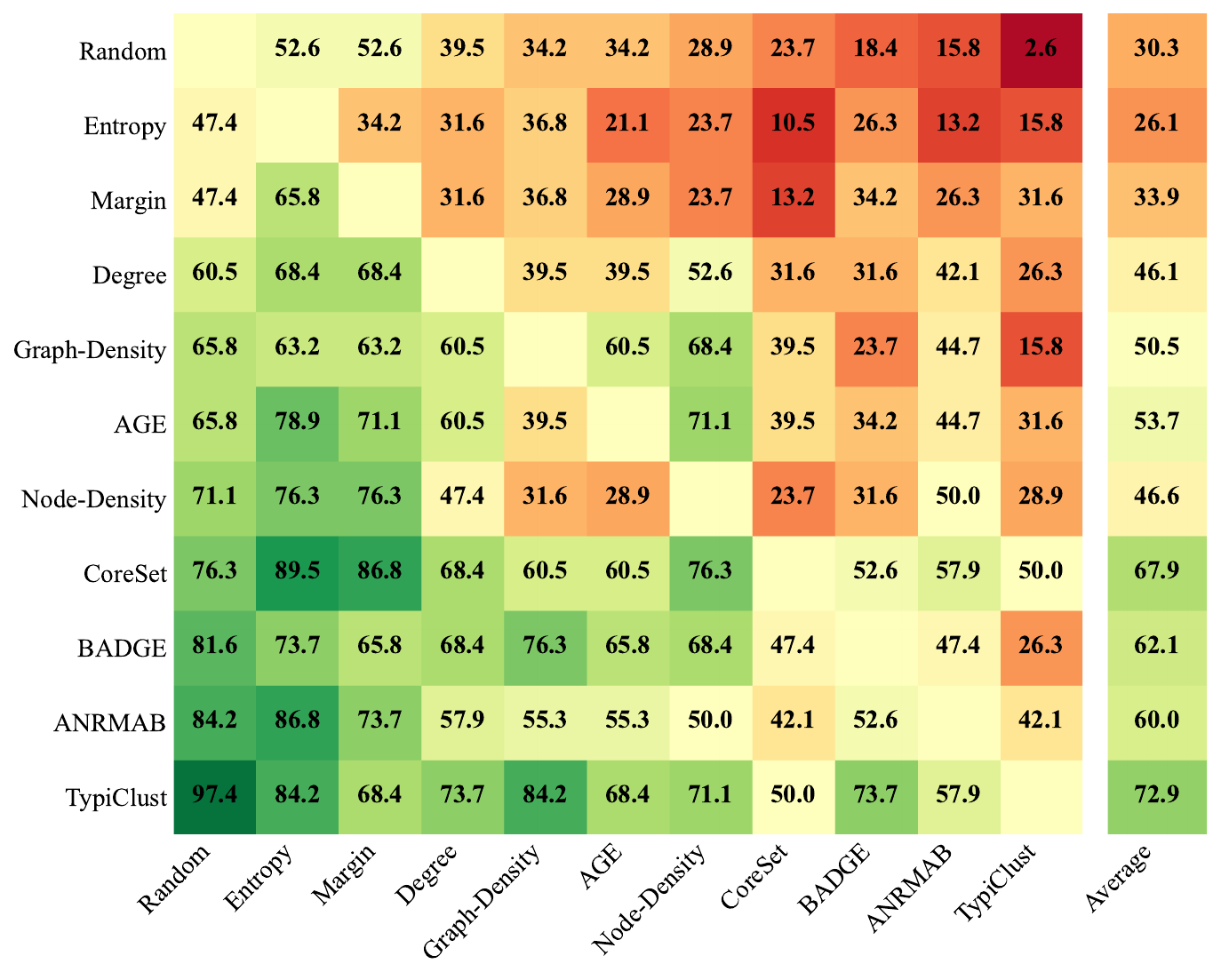}
\caption{Pairwise comparison matrix averaged across 4 datasets and 10 trials. Element~$(i, j)$ corresponds to the
proportion of total runs (in $\%$), where strategy $i$ yields a higher AULC compared to strategy $j$. For each strategy, we considered the best-performing aggregation mechanism for the respective trial.} \label{fig2}
\end{figure}

Figure~\ref{fig2} shows the pairwise comparison matrix between all strategies across all four datasets. For each trial, we considered the best-performing aggregation method for each strategy. Except for the purely uncertainty-based strategies Entropy and Margin, all strategies perform better than random sampling on average, with five strategies yielding a win rate of more than $70~\%$. The three strategies that incorporate diversity (CoreSet, BADGE, and TypiClust) exhibit the highest average win rates (row average). The best-performing strategy is TypiClust, which shows a nearly perfect win rate against random sampling and the highest average win rate against the other strategies.

\begin{figure}[t]
\includegraphics[width=\textwidth]{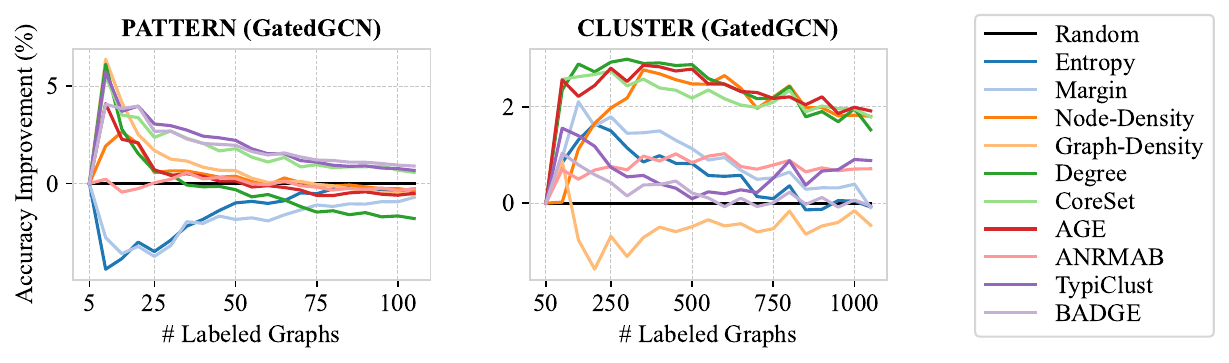}
\caption{Relative accuracy learning curves for different query strategies in the ALINC framework on PATTERN and CLUSTER using GatedGCN.} \label{fig1}
\end{figure}

The relative learning curves for GatedGCN on PATTERN and CLUSTER are presented in Fig.~\ref{fig1}. Again, we focus on comparing individual strategies by selecting the best-performing aggregation method for each strategy as measured by the AULC. We observe that the performance of most strategies varies strongly across different datasets. For example, while the purely uncertainty-based strategies, Entropy and Margin, perform poorly on PATTERN, they significantly outperform random sampling on CLUSTER. Overall, we observe the most consistent performance improvements over random sampling for TypiClust and BADGE, as they belong to the top three performing strategies on all datasets except for CLUSTER. The relative learning curves for GPS on PascalVOC-SP and COCO-SP, as well as all absolute learning curves and a tabular AULC comparison of all strategies across all datasets, are provided in Appendix~\ref{app:lcs}.

\textbf{Key Take-Away:} Among all tested strategies, TypiClust shows the highest total win rate over random sampling ($97.4~\%$). As a hybrid strategy that combines representativeness and diversity, it shows the highest consistency across datasets. In general, different datasets yield different top-performing ALINC strategies.

\subsection{Benchmarking Aggregation Methods}\label{sec:perf2}

\begin{figure}[t]
\centering
\includegraphics[width=\textwidth]{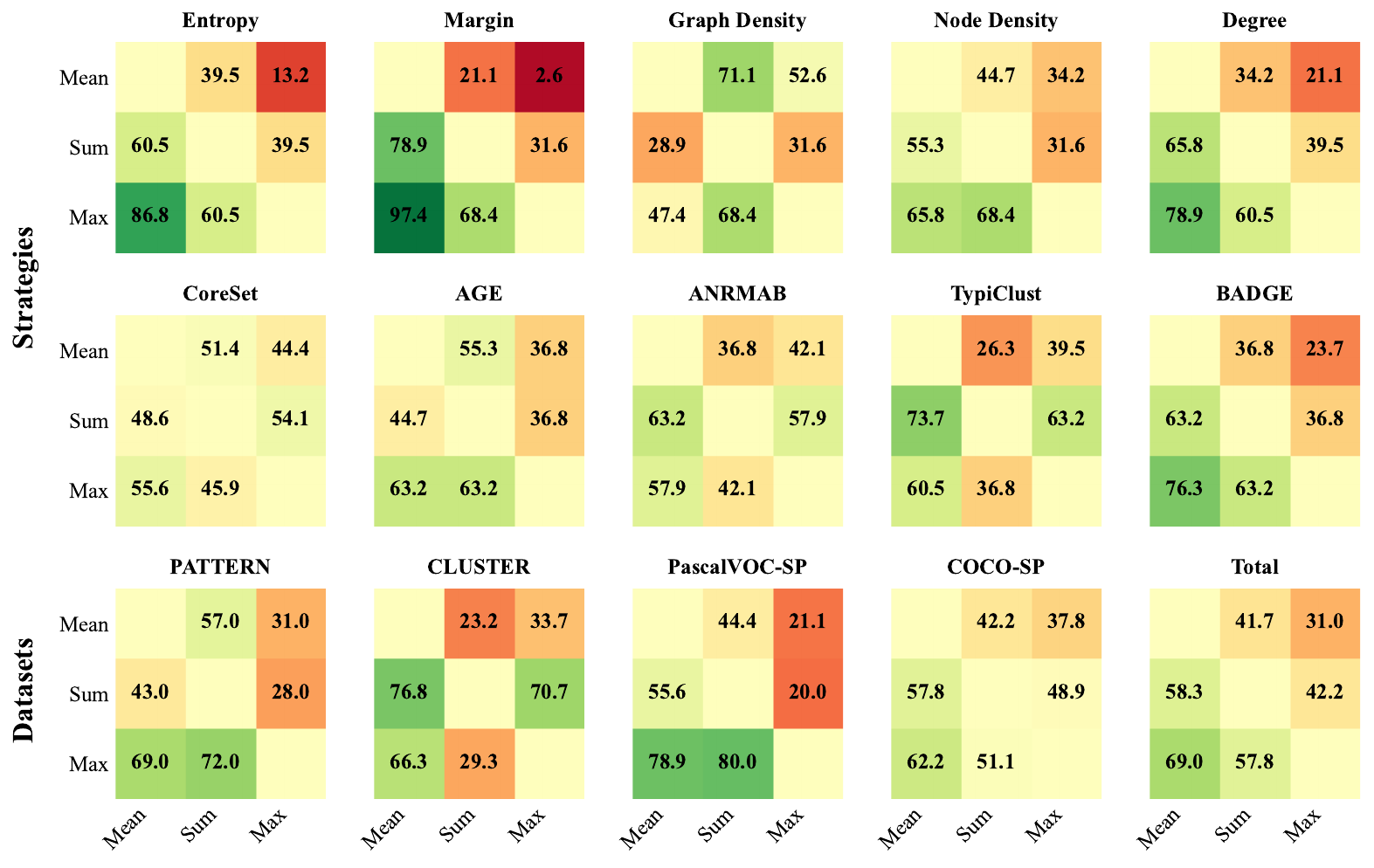}
\caption{Top and middle row: Pairwise comparison matrix averaged across 4 datasets and 10 trials for all strategies. Bottom row: Pairwise comparison matrix averaged across 10 strategies and 10 trials for all datasets, and total pairwise comparison matrix across all strategies and datasets (bottom row, rightmost column). In each subplot, element (i, j) corresponds to the proportion of total runs (in $\%$), where aggregation method i yields a higher AULC compared to aggregation method j.} \label{fig3}
\end{figure}

\begin{figure}[t]
\centering
\includegraphics[width=\textwidth]{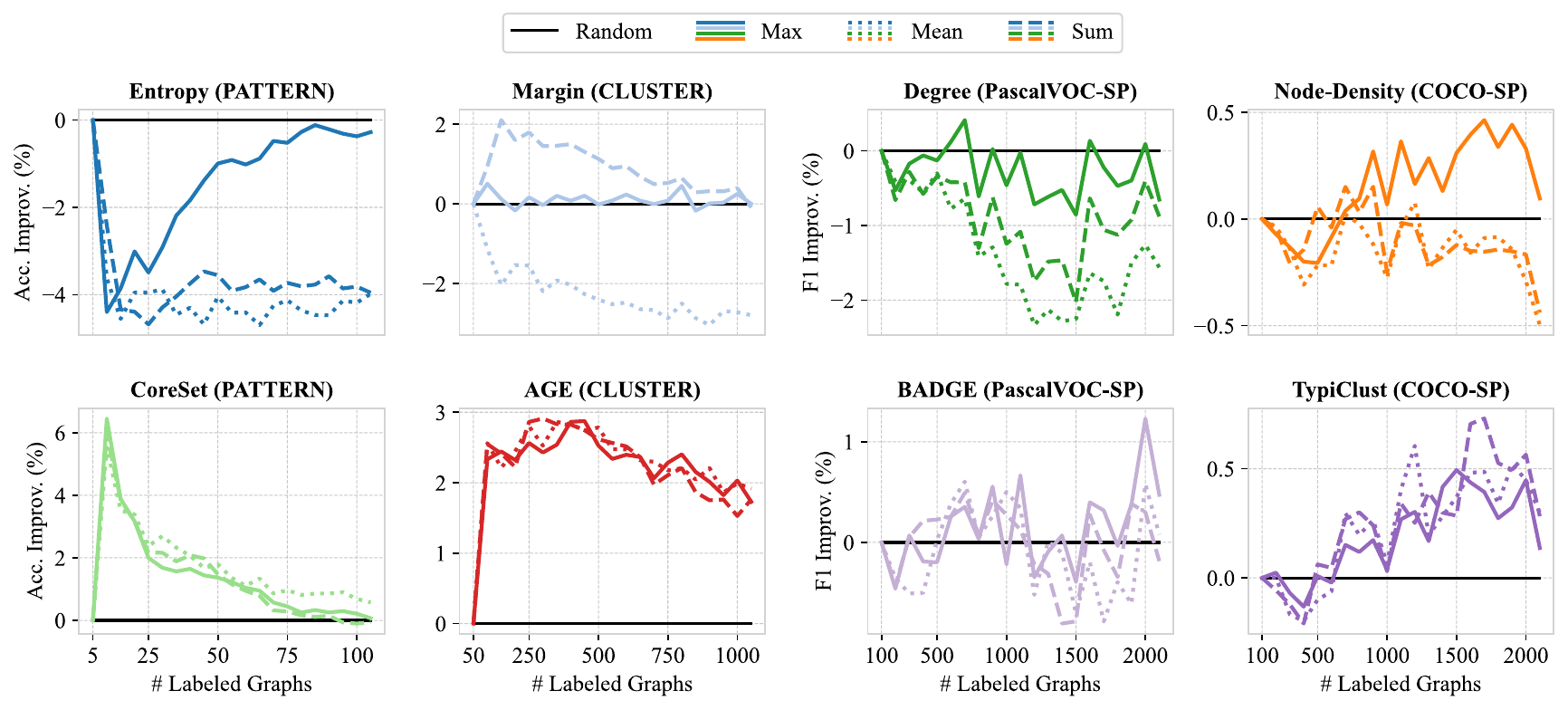}
\caption{Top row: Relative learning curves of selected strategies with high sensitivity on the aggregation type. Bottom row: Selected relative learning curves of strategies with low sensitivity on the aggregation type.} \label{fig4}
\end{figure}

Figure~\ref{fig3} shows pairwise comparisons between the three aggregation methods for all strategies and datasets included in our experiments. We observe that \texttt{mean} aggregation generally shows the lowest win rates on average, whereas \texttt{max} is the best performing aggregation method for six of ten strategies (Entropy, Margin, Node-Density, Degree, AGE, BADGE), and three of four datasets (PATTERN, PascalVOC-SP, COCO-SP). \texttt{Sum}-aggregation performs best for ANRMAB and TypiClust, as well as on the CLUSTER dataset. Among all strategies, CoreSet shows the lowest sensitivity to the choice of the aggregation function; the win rates are very balanced for this strategy. 

We further compare the relative learning curves of different aggregation mechanisms for selected strategies and datasets in Fig.~\ref{fig4}. Thereby, we observe that the impact of the aggregation method strongly depends on the strategy and the dataset. For example, the \texttt{sum}-Margin strategy outperforms random sampling on CLUSTER, whereas the \texttt{mean}-Margin strategy has a strong negative impact on model performance. For other strategies, such as CoreSet on PATTERN or TypiClust on COCO-SP, we observe rather low sensitivities regarding the aggregation type. The complete set of relative learning curves for all strategy-aggregation combinations and datasets is provided in Appendix~\ref{app:agg}.

\textbf{Key Take-Away:} The majority of ALINC strategies perform best when combined with a \texttt{max}- or \texttt{sum}-aggregation, whereas \texttt{mean}-aggregation performs worst in most cases. The choice of the aggregation method depends on the dataset and is of great importance, as it can also negatively affect the strategy, resulting in a performance worse than random sampling. 

\subsection{Annotation Cost Analysis}\label{sec:costs}

\begin{figure}[t]
\centering
\includegraphics[width=\textwidth]{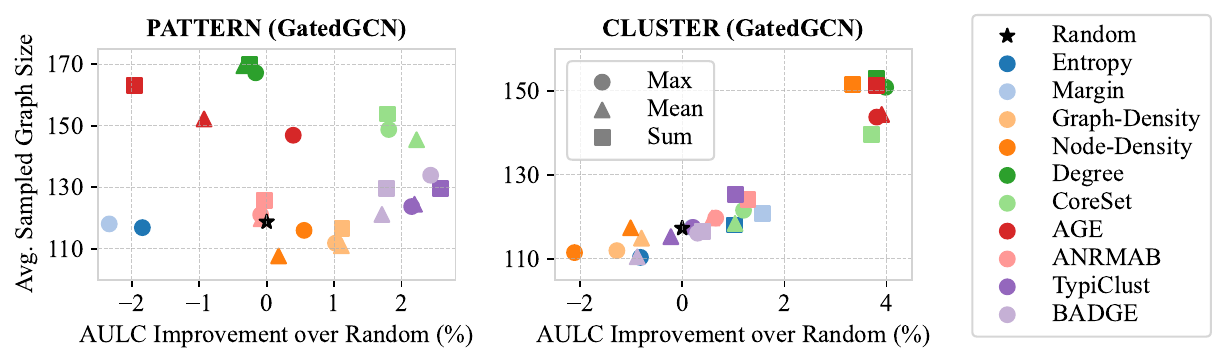}
\caption{Annotation costs in terms of averaged sampled graph size against the relative AULC improvement over random sampling for different strategies and aggregation mechanisms with GatedGCN on PATTERN and CLUSTER. Some strategy-aggregation combinations are not displayed due to poor performance.} \label{fig5}
\end{figure}

ALINC is designed for situations where the costs of labeling a single node in a graph are comparable to the costs of labeling all nodes in the graph, because both require a full-graph analysis. Therefore, the annotation costs in the ALINC framework are strongly related to the costs of the full-graph analysis, which may be influenced by different parameters of the sampled graphs, depending on the dataset or use case. For example, in some cases, the annotation costs may be proportional to the graph size (e.g., the number of nodes), whereas in other cases, the annotation costs could depend on the number of classes or the complexity of the graph structure.

For PATTERN and CLUSTER, we approximate the annotation costs of different strategy-aggregation combinations by calculating the average size of sampled graphs across all AL cycles. Figure~\ref{fig5} shows the relation between this cost measure and the AULC improvement over random sampling. We observe that some strategies, such as Degree, AGE, and CoreSet, select larger graphs on average. Furthermore, \texttt{sum}-aggregation is naturally biased towards larger graphs for many strategies. However, for the PATTERN dataset, we do not observe a significant correlation between the sampled graph size and the strategy performance. In contrast, Node- and Graph-Density sampling even achieve substantial improvements over random sampling while selecting \emph{smaller} graphs at the same time. Furthermore, TypiClust and BADGE yield even higher AULC improvements by selecting just slightly larger graphs. In contrast, on the CLUSTER dataset, we observe a strong correlation between strategy performance and the size of selected graphs, which may explain why \texttt{sum}-aggregation generally performs better on this dataset, as observed in Fig.~\ref{fig3}. 


\textbf{Key Take-Away:} While some strategies are biased towards selecting larger graphs, this is not always a guarantee for better performance. In contrast, we observed that some ALINC strategies even outperform random sampling by selecting \emph{smaller} graphs.

\subsection{Use Case Studies}\label{sec:usecase}

\begin{figure}[t]
\centering
\includegraphics[width=0.7\textwidth]{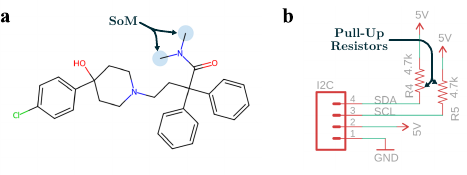}
\caption{\textbf{a)} An example molecule from the Zaretzki dataset with metabolism sites highlighted in light blue. \textbf{b)} An excerpt from a PCB schematic with two pull-up resistors (R4, R5) connecting the I2C signal lines to a 5V supply.} \label{fig6}
\end{figure}

\begin{figure}[t]
\centering
\includegraphics[width=\textwidth]{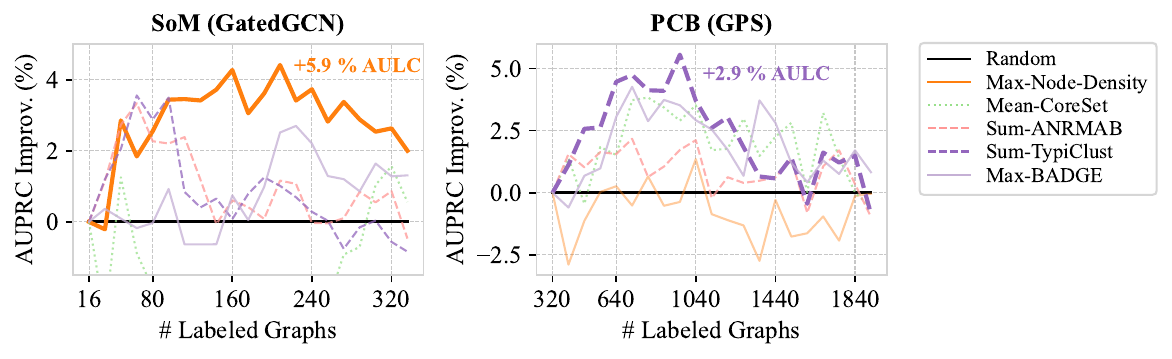}
\caption{Relative learning curves for different ALINC strategies on the Zaretzki SoM dataset using GatedGCN (left), and on the PCB pull-ups/-downs dataset using GPS (right). The strategies with the highest AULC are highlighted for better visibility. The used performance metric is the area under the precision-recall curve (AUPRC).} \label{fig7}
\end{figure}

We apply ALINC in two use case studies: site-of-metabolism (SoM) prediction in molecules, and missing pull-up/-down resistor prediction in PCB schematics.  For experimental details on both use cases, see Appendix~\ref{app:usecase}.

\subsubsection{Site-of-Metabolism Prediction.}\label{sec:som}

Predicting SoM in molecules with machine learning models is an active area of research~\cite{litsa2021machine,sicho2019fame}, but its success depends on the availability of experimentally acquired training data. While AL has been applied to the selection of specific atoms for SoM annotation based on already acquired raw measurement data~\cite{chen2024active}, we consider the use case of selecting entire molecules for experimental testing. To this end, we use the publicly available Zaretzki dataset~\cite{zaretzki2013xenosite}, which contains more than 650 drugs and drug-like compounds with experimentally confirmed SoM labels. An example compound from this dataset is visualized in Fig.~\hyperref[fig6]{\ref*{fig6}a}. We first transform the raw molecule data into graphs following the representation from the ogbg-molhiv and ogbg-molpcba datasets~\cite{hu2020open}. Using the full training set, we first perform hyperparameter optimizations on three different message-passing GNNs (GatedGCN, GINE, and GATv2), and observe that GatedGCN yields the highest validation performance (see App.~D for details). Consequently, we use GatedGCN to perform AL experiments in the setup explained in Sec.~\ref{sec:setup} with the five top-performing ALINC strategies according to our benchmark study: \texttt{max}-Node-Density, \texttt{sum}-ANRMAB, \texttt{mean}-CoreSet, \texttt{sum}-TypiClust, and \texttt{max}-BADGE. We observe that \texttt{max}-Node-Density outperforms diversity-based strategies on this dataset and yields the highest AULC with a relative improvement of $5.9~\%$ over random sampling (see Fig.~\ref{fig7}).

\subsubsection{PCB Schematic Design Automation.}\label{sec:pcb}
As a second use case, we investigate the application of ALINC to PCB schematic design automation, specifically the task of adding missing pull-up and pull-down resistors (see Fig.~\hyperref[fig6]{\ref*{fig6}b} for an example). While not necessary for circuit functionality, design optimizations like this increase the overall lifetime and reliability of the final electronic device. There is a huge interest for automating such optimization tasks, as they are usually performed manually by human engineers in a time-consuming and error-prone process. We use the dataset and graph representation from~\cite{plettenberg2025graph}, and tune hyperparameters (number of layers, hidden dimension, learning rate) via holdout validation. Thereby, we compare the same GNNs used for SoM prediction with GPS and FlowGATv2~\cite{plettenberg2025flow}, which has been shown to be effective on electronic circuit data. Since GPS yields the highest performance among the models (see App. D), we use this model for AL experiments. Note that the dataset task requires making a prediction on pairs of nodes. Therefore, we perform all aggregations on node-pair-level logits or features, which is a straightforward modification of our framework and highlights its flexibility. Figure~\ref{fig7} shows that several strategies outperform random sampling, with \texttt{sum}-TypiClust yielding the highest AULC ($+2.9~\%$ over random).

\section{Conclusion} 
We introduced ALINC, the first AL framework for inductive node classification that enables the selection of entire graphs for annotation via aggregation of node-level utility measures. Through extensive benchmarking, we identify TypiClust, CoreSet, and BADGE as top-performing strategies and reveal that the choice of the aggregation method has a strong impact on model performance and annotation cost. Successfully applied to molecular chemistry and electronic design, ALINC provides a robust, flexible solution for maximizing data efficiency in domains where node labeling requires a full-graph analysis.

In future work, we want to explore more complex aggregation methods, e.g., by introducing learnable weights. Inspired by its success in the image domain, we also want to investigate the usage of self-supervised learning to further increase the performance of ALINC strategies. Finally, we want to extend our framework also to node regression, since it would open up many new application possibilities, e.g., in GNN-based surrogate modeling for numerical simulations.

\begin{credits}
\subsubsection{\ackname} This study was funded by the  German Federal Ministry of Research, Technology, and Space (funding code 16ME0877).

\subsubsection{\discintname}
The authors have no competing interests to declare that are relevant to the content of this article.

\subsubsection{\llmusage} Large language models (LLMs) were only used for proofreading, language editing, and to assist with streamlining and debugging code. They were not used to generate scientific ideas, analyses, results, or conclusions. The authors take full responsibility for the final content of this paper.
\end{credits}
%
%
%
\bibliographystyle{splncs04}
\bibliography{references}
%
%




\newpage

\appendix
\renewcommand*{\theHsection}{app.\Alph{section}}
\renewcommand*{\theHsubsection}{\theHsection.\arabic{subsection}}

\section{Number of Clusters in Density-based Strategies}\label{app:clusters}

\begin{figure}[ht]
\centering
\includegraphics[width=\textwidth]{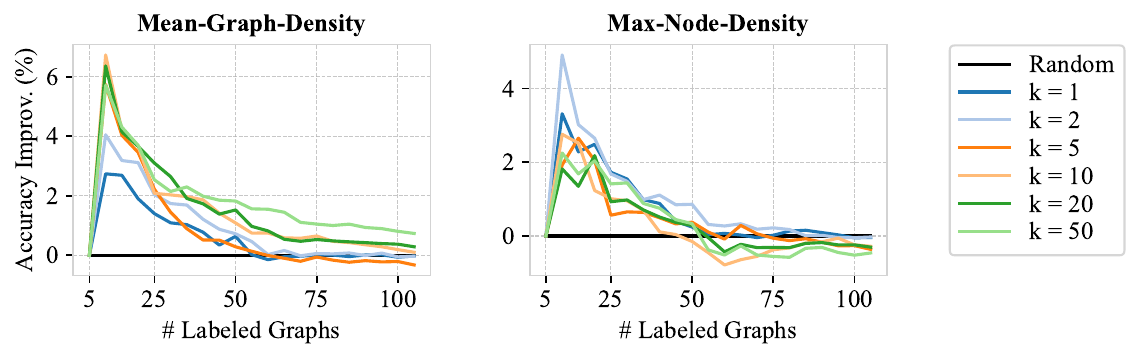}
\caption{Relative accuracy learning curves on PATTERN using GatedGCN with the strategies Mean-Graph-Density and Max-Node-Density for different values of $k$ for the $k$-\textsc{means} clustering.} \label{app:fig:clusters}
\end{figure}

Density-based acquisition strategies perform $k$-\textsc{means} clustering to calculate the information density. We investigate the influence of the number of clusters $k$ on the performance of the ALINC strategies \texttt{mean}-Graph-Density and \texttt{max}-Node-Density with GatedGCN on PATTERN (see Fig.~\ref{app:fig:clusters}). We observe that choosing $k = b = 5$ does not yield the best performance for both strategies. For \texttt{mean}-Graph-Density, the highest AULC is achieved with $k = 50$, and for \texttt{max}-Node-Density, the highest AULC is achieved with $k = 2$. For the \texttt{mean}-Graph-Density strategy, the influence of $k$ on the strategy performance is stronger than for \texttt{max}-Node-Density. 

\section{Additional Benchmark Results}\label{app:lcs}

\begin{figure}[!h]
\centering
\includegraphics[width=\textwidth]{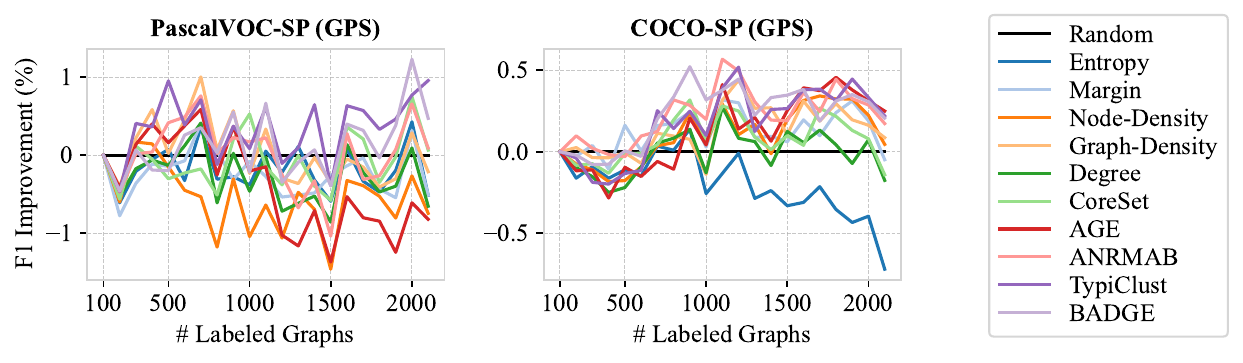}
\caption{Relative F1 learning curves for different query strategies in the ALINC framework on PascalVOC-SP and COCO-SP using GPS.} \label{app:fig:rel_sp}
\end{figure}

\begin{figure}[!h]
\centering
\includegraphics[width=\textwidth]{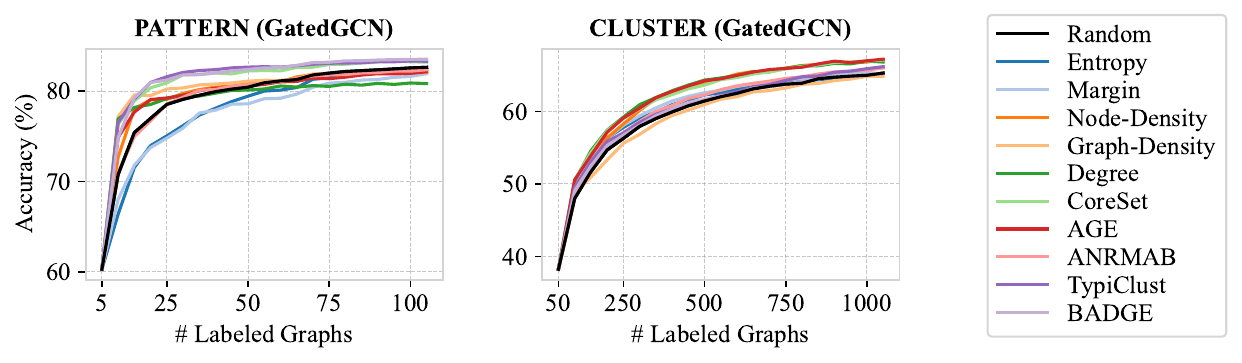}
\caption{Absolute accuracy learning curves for different query strategies in the ALINC framework on PATTERN and CLUSTER using GatedGCN.} \label{app:fig:abs_sbm}
\end{figure}

\begin{figure}[!h]
\centering
\includegraphics[width=\textwidth]{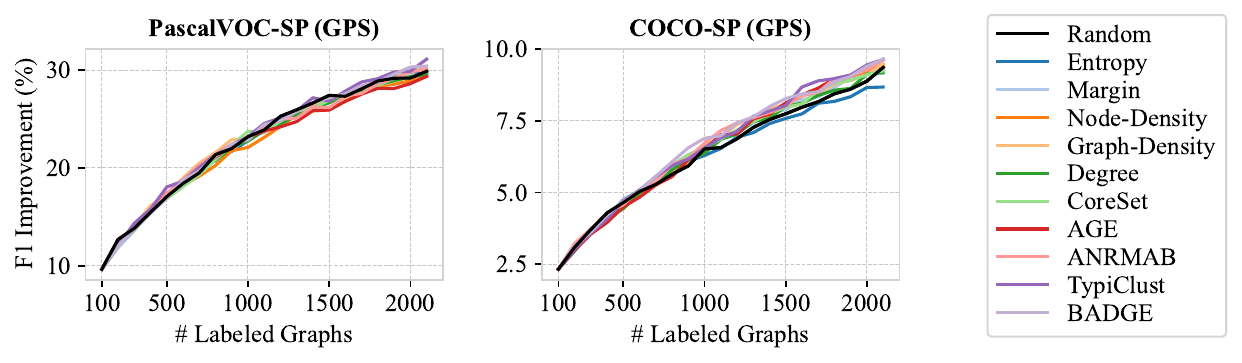}
\caption{Absolute F1 learning curves for different query strategies in the ALINC framework on PascalVOC-SP and COCO-SP using GPS.} \label{app:fig:abs_sp}
\end{figure}

Supplementing the relative learning curves provided in the main paper, we also provide the relative learning curves for all strategies (with the best aggregation method, respectively) on PascalVOC-SP and COCO-SP in Fig.~\ref{app:fig:rel_sp}. Furthermore, absolute learning curves are provided in Fig.~\ref{app:fig:abs_sbm} (PATTERN and CLUSTER) and Fig.~\ref{app:fig:abs_sp} (PascalVOC-SP and COCO-SP).

Finally, the AULC results for all strategy-aggregation combinations on all four datasets, averaged over 10 runs with different random seeds, are provided in Tab.~\ref{app:tab:aulc}.

\clearpage

\begin{table}[!h]
\caption{AULC (in $\%$) for all strategy-aggregation comparisons and all datasets, averaged across 10 runs with different random seeds, respectively. The best strategy for each dataset is highlighted in bold.\label{app:tab:aulc}}
\scriptsize
\renewcommand{\arraystretch}{1.5}
\begin{tabularx}{\textwidth}{
    >{\hsize=1.1\hsize}X 
    >{\hsize=0.7\hsize}X|
    >{\hsize=1.05\hsize}R 
    >{\hsize=1.05\hsize}R 
    >{\hsize=1.05\hsize}R 
    >{\hsize=1.05\hsize}R 
}
\hline
\textbf{Strategy} & \textbf{Aggreg.} & \textbf{PATTERN} & \textbf{CLUSTER} & \textbf{VOC-SP} & \textbf{COCO-SP} \\
\hline
Random & - & 79.57 $\pm$ 0.97 & 59.97 $\pm$ 0.20 & 22.77 $\pm$ 0.46 & 6.45 $\pm$ 0.27 \\ \hline

\multirow{3}{*}{Entropy} & Mean & 75.42 $\pm$ 1.98 & 57.47 $\pm$ 0.24 & 21.76 $\pm$ 0.64 & 6.26 $\pm$ 0.15 \\
                         & Sum  & 75.83 $\pm$ 1.40 & 60.57 $\pm$ 0.22 & 21.64 $\pm$ 0.53 & 6.17 $\pm$ 0.18 \\
                         & Max  & 78.10 $\pm$ 1.77 & 59.48 $\pm$ 0.26 & 22.50 $\pm$ 0.48 & 6.17 $\pm$ 0.25 \\ \hline

\multirow{3}{*}{Margin}  & Mean & 75.23 $\pm$ 2.52 & 57.70 $\pm$ 0.32 & 21.57 $\pm$ 0.39 & 6.04 $\pm$ 0.24 \\
                         & Sum  & 76.46 $\pm$ 1.67 & 60.91 $\pm$ 0.16 & 21.59 $\pm$ 0.44 & 6.16 $\pm$ 0.27 \\
                         & Max  & 77.70 $\pm$ 1.91 & 60.08 $\pm$ 0.13 & 22.67 $\pm$ 0.76 & 6.58 $\pm$ 0.21 \\ \hline

\multirow{3}{*}{Node-Density} & Mean & 79.71 $\pm$ 0.62 & 59.37 $\pm$ 0.61 & 20.89 $\pm$ 0.42 & 6.27 $\pm$ 0.25 \\
                              & Sum  & 77.42 $\pm$ 1.14 & 61.98 $\pm$ 0.28 & 21.41 $\pm$ 0.62 & 6.32 $\pm$ 0.27 \\
                              & Max  & 80.01 $\pm$ 0.53 & 58.70 $\pm$ 0.44 & 22.32 $\pm$ 0.59 & 6.55 $\pm$ 0.18 \\ \hline

\multirow{3}{*}{Graph-Density} & Mean & 80.45 $\pm$ 0.46 & 59.48 $\pm$ 0.28 & 22.82 $\pm$ 0.40 & 6.42 $\pm$ 0.14 \\
                               & Sum  & 80.46 $\pm$ 0.81 & 56.96 $\pm$ 0.40 & 22.70 $\pm$ 0.36 & 6.46 $\pm$ 0.21 \\
                               & Max  & 80.38 $\pm$ 0.86 & 59.19 $\pm$ 0.38 & 22.74 $\pm$ 0.45 & 6.58 $\pm$ 0.27 \\ \hline

\multirow{3}{*}{Degree}  & Mean & 79.30 $\pm$ 1.45 & 62.25 $\pm$ 0.27 & 21.48 $\pm$ 0.49 & 5.40 $\pm$ 0.18 \\
                         & Sum  & 79.37 $\pm$ 1.34 & 62.26 $\pm$ 0.29 & 21.92 $\pm$ 0.33 & 6.45 $\pm$ 0.10 \\
                         & Max  & 79.44 $\pm$ 1.47 & \textbf{62.36 $\pm$ 0.26} & 22.57 $\pm$ 0.59 & 5.95 $\pm$ 0.22 \\ \hline

\multirow{3}{*}{CoreSet} & Mean & 81.34 $\pm$ 0.69 & 60.58 $\pm$ 0.34 & 22.14 $\pm$ 0.45 & 6.54 $\pm$ 0.16 \\
                         & Sum  & 80.99 $\pm$ 0.68 & 62.19 $\pm$ 0.28 & 22.17 $\pm$ 0.48 & 6.41 $\pm$ 0.35 \\
                         & Max  & 81.01 $\pm$ 1.07 & 60.68 $\pm$ 0.32 & 22.69 $\pm$ 0.45 & 6.44 $\pm$ 0.13 \\ \hline

\multirow{3}{*}{AGE}     & Mean & 78.82 $\pm$ 1.10 & 62.31 $\pm$ 0.19 & 21.47 $\pm$ 0.57 & 6.47 $\pm$ 0.24 \\
                         & Sum  & 78.01 $\pm$ 1.82 & 62.25 $\pm$ 0.25 & 21.86 $\pm$ 0.58 & 6.57 $\pm$ 0.18 \\
                         & Max  & 79.89 $\pm$ 1.03 & 62.26 $\pm$ 0.16 & 22.42 $\pm$ 0.57 & 6.42 $\pm$ 0.27 \\ \hline

\multirow{3}{*}{ANRMAB}  & Mean & 79.50 $\pm$ 1.14 & 60.32 $\pm$ 0.16 & 22.45 $\pm$ 0.53 & 6.57 $\pm$ 0.23 \\
                         & Sum  & 79.53 $\pm$ 0.65 & 60.74 $\pm$ 0.27 & 22.53 $\pm$ 0.42 & 6.67 $\pm$ 0.24 \\
                         & Max  & 79.49 $\pm$ 0.82 & 60.36 $\pm$ 0.16 & 22.77 $\pm$ 0.56 & \textbf{6.68 $\pm$ 0.13} \\ \hline

\multirow{3}{*}{TypiClust} & Mean & 81.30 $\pm$ 0.44 & 59.84 $\pm$ 0.30 & 22.84 $\pm$ 0.57 & 6.63 $\pm$ 0.24 \\
                           & Sum  & \textbf{81.62 $\pm$ 0.49} & 60.58 $\pm$ 0.20 & 22.58 $\pm$ 0.37 & 6.63 $\pm$ 0.23 \\
                           & Max  & 81.28 $\pm$ 0.55 & 60.09 $\pm$ 0.29 & \textbf{23.11 $\pm$ 0.33} & 6.57 $\pm$ 0.17 \\ \hline

\multirow{3}{*}{BADGE}   & Mean & 80.93 $\pm$ 0.56 & 59.44 $\pm$ 0.15 & 22.67 $\pm$ 0.31 & 6.65 $\pm$ 0.16 \\
                         & Sum  & 80.99 $\pm$ 0.41 & 60.21 $\pm$ 0.19 & 22.77 $\pm$ 0.54 & \textbf{6.68 $\pm$ 0.18} \\
                         & Max  & 81.50 $\pm$ 0.29 & 60.15 $\pm$ 0.29 & 22.90 $\pm$ 0.40 & 6.52 $\pm$ 0.20 \\ \hline
\end{tabularx}
\end{table}

\clearpage
\section{Additional Aggregation Comparisons}\label{app:agg}

\begin{figure}[!h]
\centering
\includegraphics[width=\textwidth]{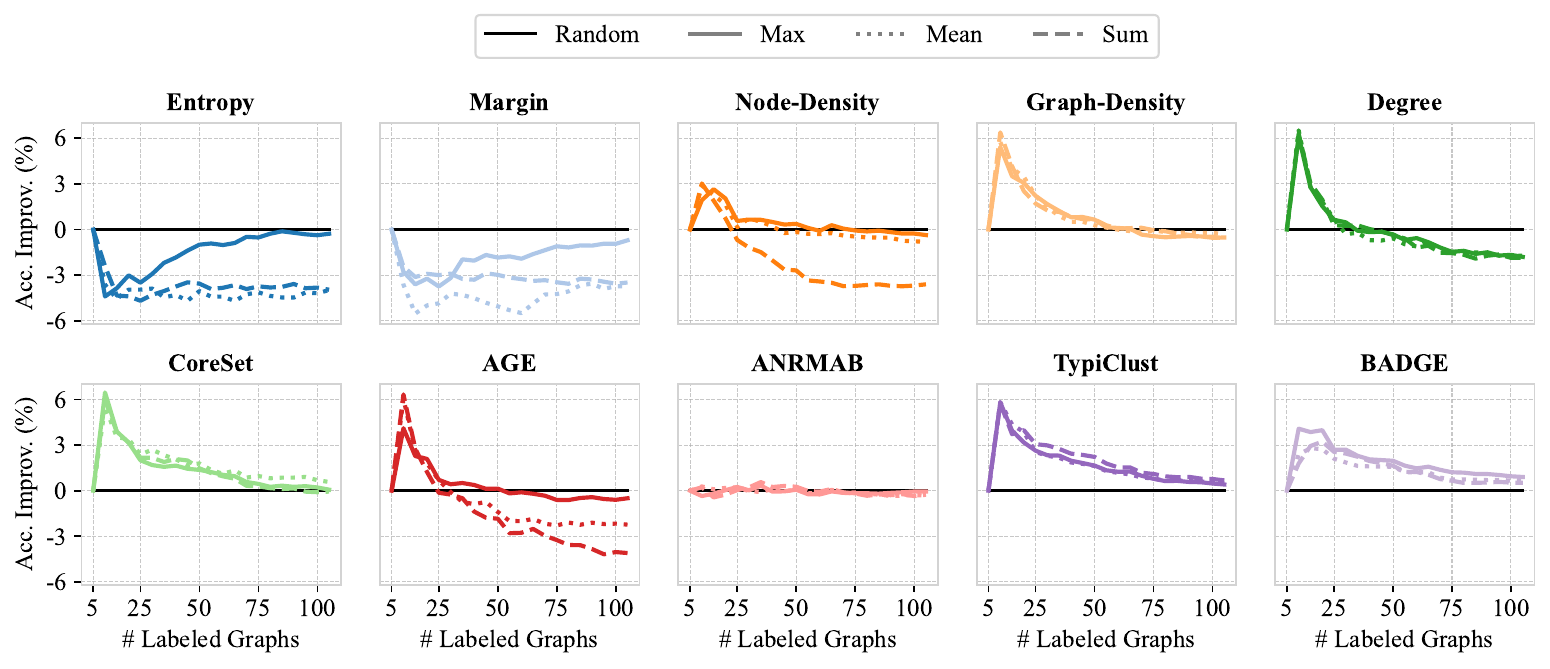}
\caption{Relative accuracy learning curves for all strategies with different aggregation methods on PATTERN with GatedGCN.} \label{app:fig:agg_pattern}
\end{figure}

\begin{figure}[!h]
\centering
\includegraphics[width=\textwidth]{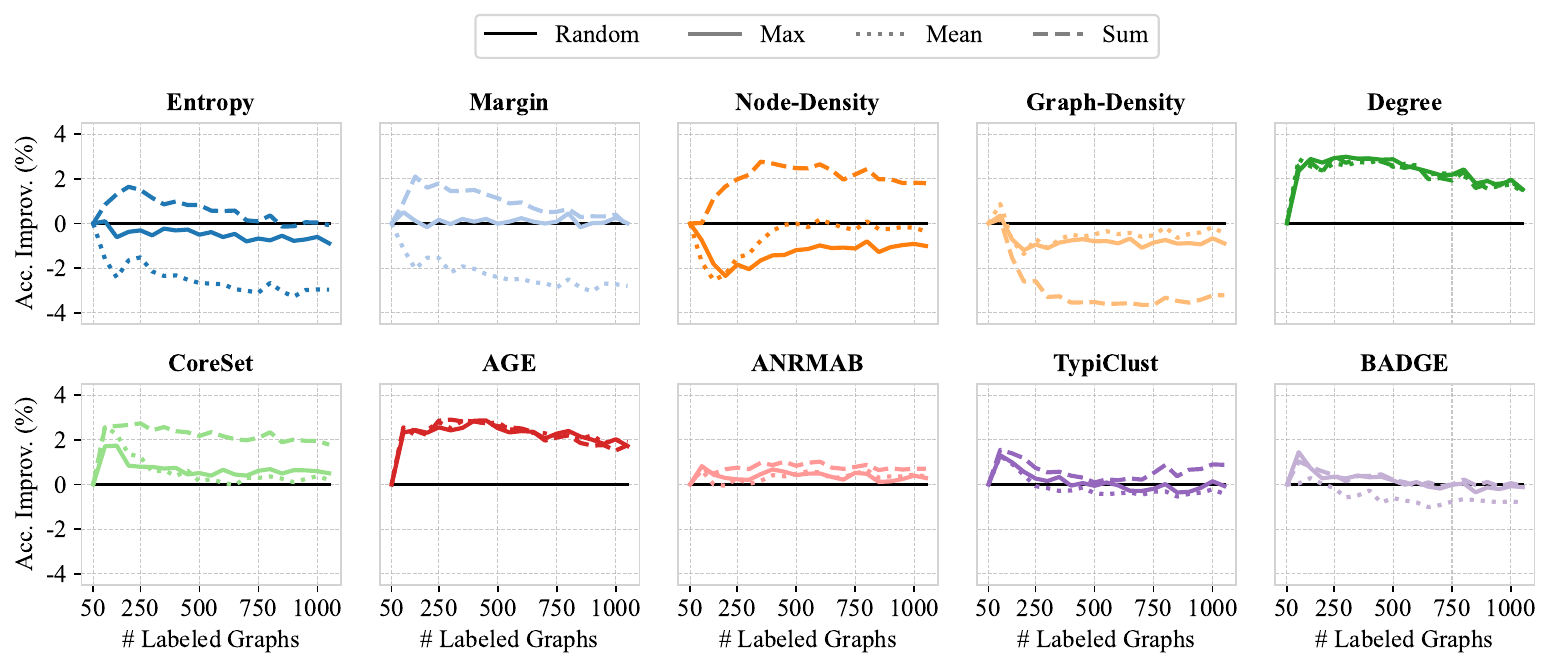}
\caption{Relative accuracy learning curves for all strategies with different aggregation methods on CLUSTER with GatedGCN.} \label{app:fig:agg_cluster}
\end{figure}

\begin{figure}[!h]
\centering
\includegraphics[width=\textwidth]{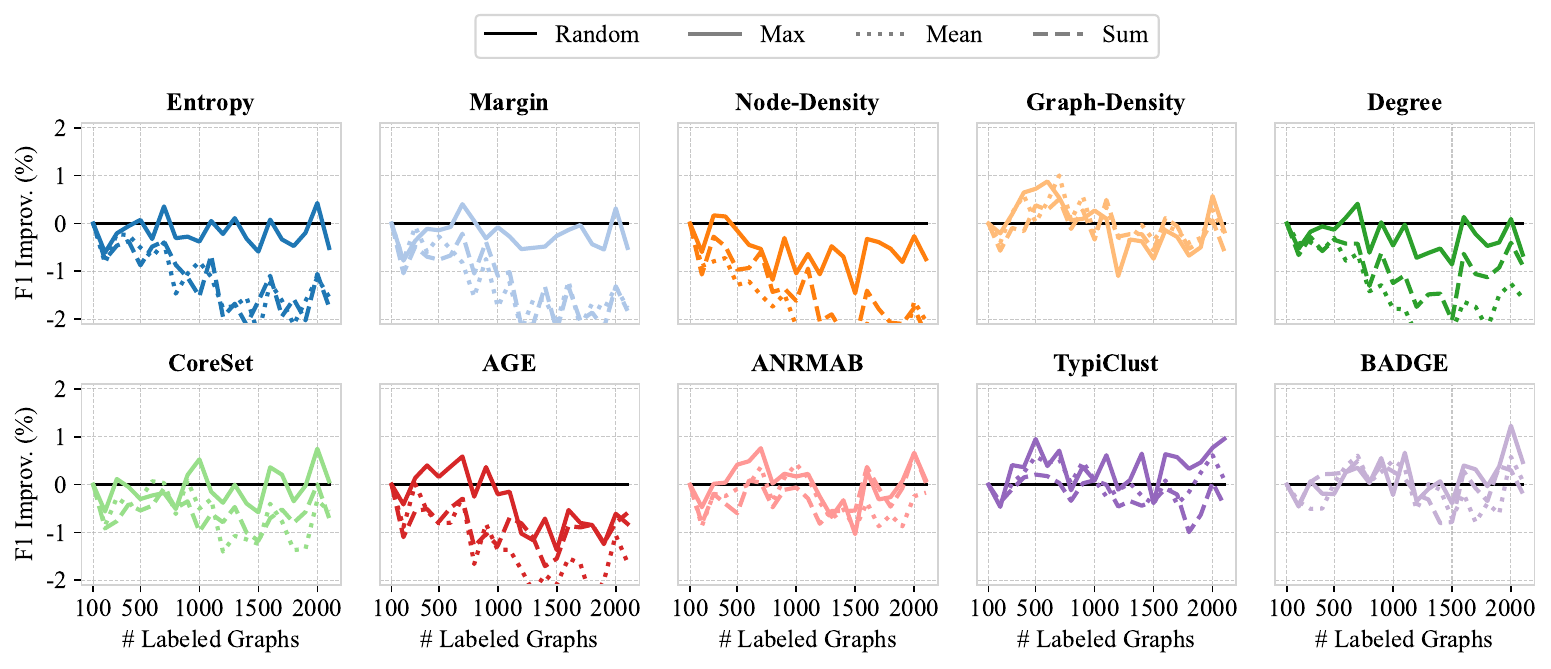}
\caption{Relative F1 learning curves for all strategies with different aggregation methods on PascalVOC-SP with GPS.} \label{app:fig:agg_pascal}
\end{figure}

\begin{figure}[!h]
\centering
\includegraphics[width=\textwidth]{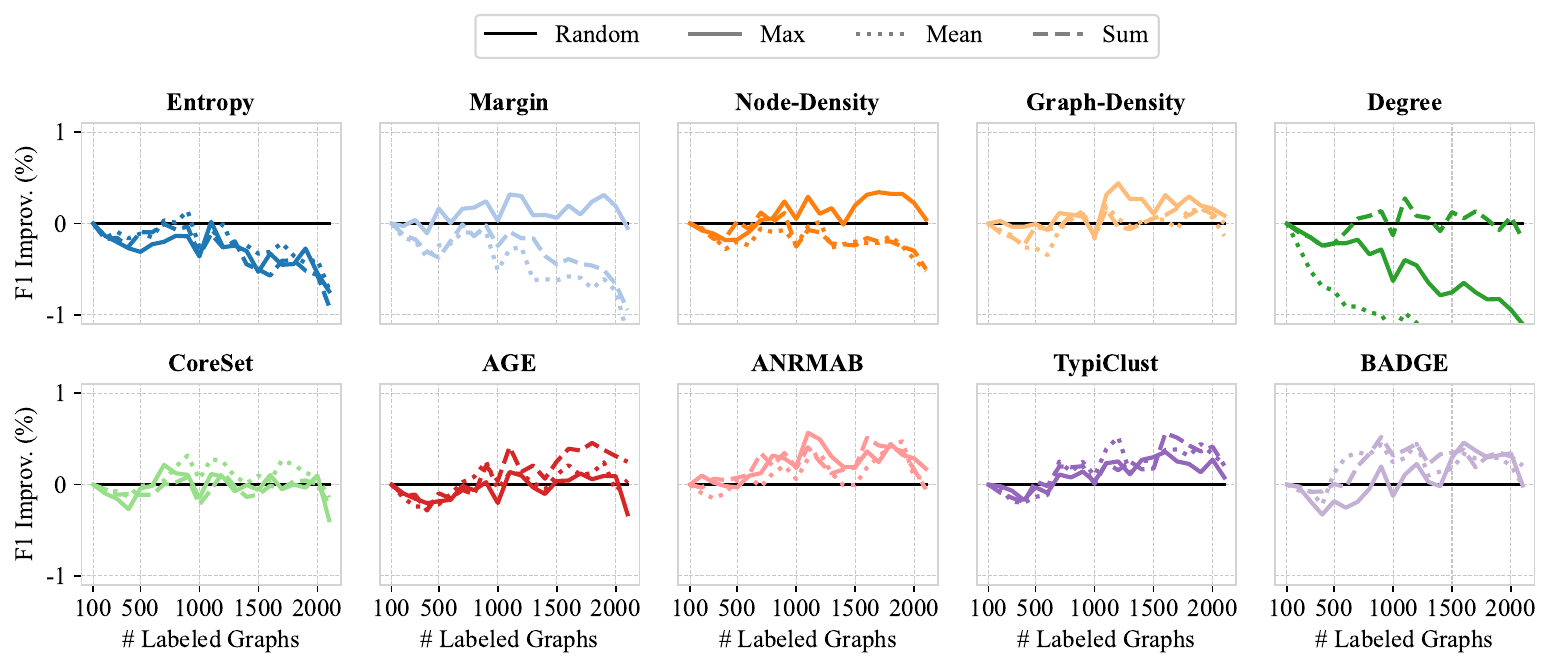}
\caption{Relative F1 learning curves for all strategies with different aggregation methods on COCO-SP with GPS.} \label{app:fig:agg_coco}
\end{figure}

Aggregation method comparisons were provided for selected strategies and datasets in Fig.~3 of the main paper. Here, we present similar comparisons for all strategies and datasets in Fig.~\ref{app:fig:agg_pattern} (PATTERN), Fig.~\ref{app:fig:agg_cluster} (CLUSTER), Fig.~\ref{app:fig:agg_pascal} (PascalVOC-SP), and Fig.~\ref{app:fig:agg_coco} (COCO-SP).

\clearpage
\section{Experimental Details on the Use Case Studies}\label{app:usecase}

In this section, we provide further details on the use case studies presented in Sec.~\ref{sec:usecase} of the main paper.

\subsection{Site-of-Metabolism Prediction}
We utilize the Zaretzki dataset~\cite{zaretzki2013xenosite} and convert the molecules into graphs using the representation proposed for the ogbg-molhiv and ogbg-molpcba datasets from~\cite{hu2020open}. Each molecular graph further obtains binary node labels corresponding to the SoMs. The final graph dataset is summarized in Tab.~\ref{app:tab:data}.

\begin{table}[!h]
\caption{\label{app:tab:data} Overview of the graph datasets for the use case studies.}
\scriptsize
\renewcommand{\arraystretch}{1.5}
\begin{tabularx}{\textwidth}{
    >{\hsize=1.0\hsize}X| 
    >{\hsize=0.9\hsize}R 
    >{\hsize=0.9\hsize}R 
    >{\hsize=0.9\hsize}R 
    >{\hsize=1.1\hsize}R 
    >{\hsize=1.1\hsize}R 
}
\hline
\textbf{Dataset} & \textbf{\#Graphs} & \textbf{Avg. Nodes} & \textbf{Avg. Edges} & \textbf{\#Classes (C)} & \textbf{Acq. Size (b)} \\
\hline
Zaretzki          & 657             &        22.65          & 48.87               & 2                   & 16                   \\
PCB-Pull-Ups/-Downs        & 2406            & 139.84                & 382.94               & 2                   & 80                  \\
\hline
\end{tabularx}
\end{table}

We perform a scaffold split \cite{bemis1996properties} to divide the dataset into train, validation, and test splits using a ratio of $80/10/10~\%$. Afterwards, we train three different GNN models (GINE~\cite{hu2020strategies}, GATv2~\cite{brody2022attentive}, and GatedGCN~\cite{bresson2017residual,dwivedi2023benchmarking}) on the full training set and use the holdout validation set to perform hyperparameter optimizations. Thereby, we tune the learning rate and the number of message-passing layers using a simple grid search. We choose the hidden dimension according to a fixed parameter budget of roughly $500,000$. The test set performances of the optimized GNN models are reported in Tab.~\ref{app:tab:zaretzki}. Besides the area under the precision-recall curve (AUPRC), which we used as the main evaluation metric, we also provide the top-2 and top-3 metrics, as well as the Matthew correlation coefficient (MCC), following the evaluation scheme from~\cite{sicho2019fame}. GatedGCN shows the best test set performance when trained on the full training set and is therefore selected for the AL experiments. We provide the full set of hyperparameters for all models in our code repository\footnote{\href{https://github.com/pasplett/alinc}{https://github.com/pasplett/alinc}.}

\begin{table}[!ht]
\caption{\label{app:tab:zaretzki} Test set performance of different message-passing GNNs trained on the Zaretzki dataset.}
\scriptsize
\renewcommand{\arraystretch}{1.5}
\begin{tabularx}{\textwidth}{
    >{\hsize=0.8\hsize}X|
    >{\hsize=1.05\hsize}R
    >{\hsize=1.05\hsize}R 
    >{\hsize=1.05\hsize}R 
    >{\hsize=1.05\hsize}R 
}
\hline
Model    & AUPRC ($\%, \uparrow$)  & Top-2 ($\%, \uparrow$)  & Top-3 ($\%, \uparrow$)  & MCC ($\uparrow$) \\
\hline
GINE     & $58.1 \pm 2.1$          & $81.2 \pm 3.1$          & $88.5 \pm 2.1$          & $0.53 \pm 0.01$  \\
GATv2    & $61.3 \pm 1.3$          & $77.9 \pm 3.0$          & $85.2 \pm 1.8$          & $0.53 \pm 0.01$  \\
GatedGCN & $\mathbf{61.6 \pm 2.5}$ & $\mathbf{84.5 \pm 2.2}$ & $\mathbf{89.4 \pm 1.4}$ & $\mathbf{0.56 \pm 0.02}$ \\
\hline
\end{tabularx}
\end{table}

For the AL experiments, we stick to the AL setting described in Sec.~\ref{sec:setup} of the main paper. The acquisition size $b$ is set to 16.

\subsection{PCB Schematics Design Automation}

Wu use a proprietary PCB design automation dataset introduced in~\cite{plettenberg2025graph} for the prediction of missing pull-up and pull-down resistors in real-world PCB schematics. The schematics are represented as bipartite graphs consisting of two types of nodes (symbol nodes, such as resistors or capacitors; and net nodes, such as ground or supply nets) connected by edges representing the pins (the terminals of the symbols). All nodes and pins are associated with a reference name (such as \textit{R\$10}). These names are non-standardized and therefore processed as text attributes rather than categorical features. The popular, lightweight text encoder All-MiniLM-L6-v2~\cite{wang2020minilm} is used to generate numerical node and edge feature vectors from the raw text attributes. Further details on the dataset construction can be found in~\cite{plettenberg2025graph}. The dataset statistics are provided in Fig.~\ref{app:tab:data}.

The task for this dataset is to identify pairs of net nodes in the graph that serve as connection points for missing pull-up or pull-down resistors. Since this task is not a classical node classification but rather a \textit{node pair classification} task, we adapt the ALINC framework accordingly: Instead of evaluating the different acquisition functions based on node-level logits $\mathbf{z}^{(i)}_j$ or embeddings $\mathbf{h}^{(i)}_j$, we utilize node-pair level logits $\mathbf{z}^{(i)}_{jk}$ (obtained from the model architecture described in~\cite{plettenberg2025graph}) and node-pair level embeddings $\mathbf{h}^{(i)}_{jk} = \mathbf{h}^{(i)}_{j}~||~\mathbf{h}^{(i)}_{k}$.

Note that the model for solving the PCB design automation task from~\cite{plettenberg2025graph} uses two classification heads: First, a pre-filter classification is performed on the single nodes to identify promising candidates for possible connection points. The actual node-pair classification is then only performed on all net nodes, where the predicted probability of the pre-filter is higher than a fixed threshold $\lambda$, which we set to $0.1$. This model architecture increases the efficiency and training stability for the node pair classification, but also has an important consequence for the ALINC framework. Since the performance of the node-pair classification relies on a good pre-filtering, it performs rather poorly when trained on very small datasets. Therefore, we increase the number of initial, randomly sampled, labeled graphs to $4 \cdot b$ for this dataset. In our experiments, we choose $b = 80$.

\begin{table}[!ht]
\caption{\label{app:tab:pcb} Test set performance of different message-passing GNNs (GINE, GATv2, GatedGCN, FlowGATv2) and one GT (GPS) trained on the PCB-Pull-Ups/-Downs dataset.}
\scriptsize
\centering
\renewcommand{\arraystretch}{1.5}
\begin{tabularx}{0.6\textwidth}{
    >{\hsize=0.9\hsize}X|
    >{\hsize=1.05\hsize}R
    >{\hsize=1.05\hsize}R 
}
\hline
Model     & AUPRC ($\%, \uparrow$)  & F1 ($\%, \uparrow$)     \\
\hline
GINE      & $83.8 \pm 3.6$          & $85.3 \pm 1.9$          \\
GATv2     & $84.5 \pm 1.7$          & $84.5 \pm 1.7$          \\
FlowGATv2 & $85.4 \pm 2.6$          & $83.8 \pm 2.3$          \\
GatedGCN  & $88.4 \pm 1.7$          & $88.3 \pm 2.3$          \\
GPS       & $\mathbf{91.8 \pm 1.0}$ & $\mathbf{94.2 \pm 1.0}$ \\
\hline
\end{tabularx}
\end{table}
\FloatBarrier

Similar to the SoM use case, we first identify the best performing model on the full dataset before applying it in our ALINC setting. Additionally to the models tested on the Zaretzki dataset, we also consider FlowGATv2~\cite{plettenberg2025flow} and GPS~\cite{rampavsek2022recipe}. On all of these models, we perform a hyperparameter optimization using a simple grid search to determine the learning rate and the number of layers at a fixed parameter budget of $500,000$. Thereby, we optimize the model performance on a holdout validation set. The final test set performances of the optimized models in terms of the AUPRC and the F1-score are presented in Tab.~\ref{app:tab:pcb}. GPS significantly outperforms all other models, which is why we selected this model for the AL experiments.

\end{document}